\documentclass[10pt]{article}

\usepackage{supra3d}

\usepackage{amsmath,amsfonts,bm}

\def\eqref#1{equation~\ref{#1}}

\def\1{\bm{1}}

\DeclareMathAlphabet{\mathsfit}{\encodingdefault}{\sfdefault}{m}{sl}
\SetMathAlphabet{\mathsfit}{bold}{\encodingdefault}{\sfdefault}{bx}{n}

\usepackage{hyperref}
\usepackage{url}
\usepackage{booktabs}
\usepackage{array}
\usepackage{graphicx}
\usepackage{xcolor}
\usepackage{amsmath,amssymb}
\usepackage{pifont}

\usepackage{xspace}
\usepackage{placeins}
\DeclareCaptionFont{manuscriptcaption}{\fontsize{9.5}{11.5}\selectfont}
\newcommand{\methodname}{Gate3DGen\xspace}
\newcommand{\eg}{e.g.\@\xspace}

\usepackage{marvosym}

\newcommand{\correspondingmark}{\textsuperscript{\supracorresponding}}

\newcommand{\xo}{x^{o}}                 
\newcommand{\xc}{x^{c}}                 
\newcommand{\xchat}{\widehat{x}^{c}}    
\newcommand{\enc}{\mathcal{E}}          
\newcommand{\dec}{\mathcal{D}}          
\newcommand{\gen}{\mathcal{G}}          
\newcommand{\gate}{\mathcal{R}_{\phi}}  
\newcommand{\canon}{\mathcal{C}_{\theta}} 

\newcommand{\NumAssets}{3,961}
\newcommand{\NumTrainPairs}{43,936}
\newcommand{\NumCanonicalViews}{eight}
\newcommand{\NumViews}{four}

\newcommand{\NumEvalCases}{1,082}

\newcommand{\NumGateDegraded}{1,082}
\newcommand{\NumGateClean}{934}
\newcommand{\NumWildCases}{96}

\title{Gate3DGen:}
\suprasubtitle{A Pre-processing Model that Selectively Purifies\\Real-World Images for 3D Generation}
\suprashorttitle{Gate3DGen}
\hypersetup{
  colorlinks=true,pdfborder={0 0 0},
  pdftitle={Gate3DGen: A Pre-processing Model that Selectively Purifies Real-World Images for 3D Generation},
  pdfauthor={Xiaolin Li, Zihao Huang, Fanpeng Meng, Chen Yang, Guanjun Wu, Kai Zhang, Jiemin Fang, Qi Tian}
}
\supraabstract{    Single-image 3D generation has achieved high fidelity on curated benchmarks, yet degrades sharply on real-world images with uneven illumination, blur, and occlusions.
    Purifying such inputs requires deciding when to intervene and preserving the visual cues needed for 3D generation, while leaving clean inputs unchanged.
    We introduce \textcolor{supraAccent}{\textbf{\methodname}}, a selective pre-processing model that gates image inputs to existing 3D generators without retraining them.
    A degradation-aware router bypasses inputs classified as clean and extracts features to guide purification of the remaining inputs.
    Conditioned on these features, a 3D-oriented one-step purifier purifies degraded inputs with spatial losses that encourage preservation of object identity, contours, and textural cues.
    To provide paired supervision, we construct a synthetic data engine that renders clean views of 3D assets and synthesises composite degradations.
    On the synthetic benchmark, the purifier improves image fidelity and downstream 3D geometry in only $2.3$\,s on average, reducing TRELLIS.2 mesh distance by $61.5\%$ relative to degraded inputs.
    VLM evaluation and human preferences on real photographs further support identity preservation and degradation correction.
    More demos and code are available at
\href{https://supra-3d.github.io/gate3dgen}{\texttt{supra-3d.github.io/gate3dgen}}.

}
\suprateaser{%
  \includegraphics[width=\linewidth,keepaspectratio]{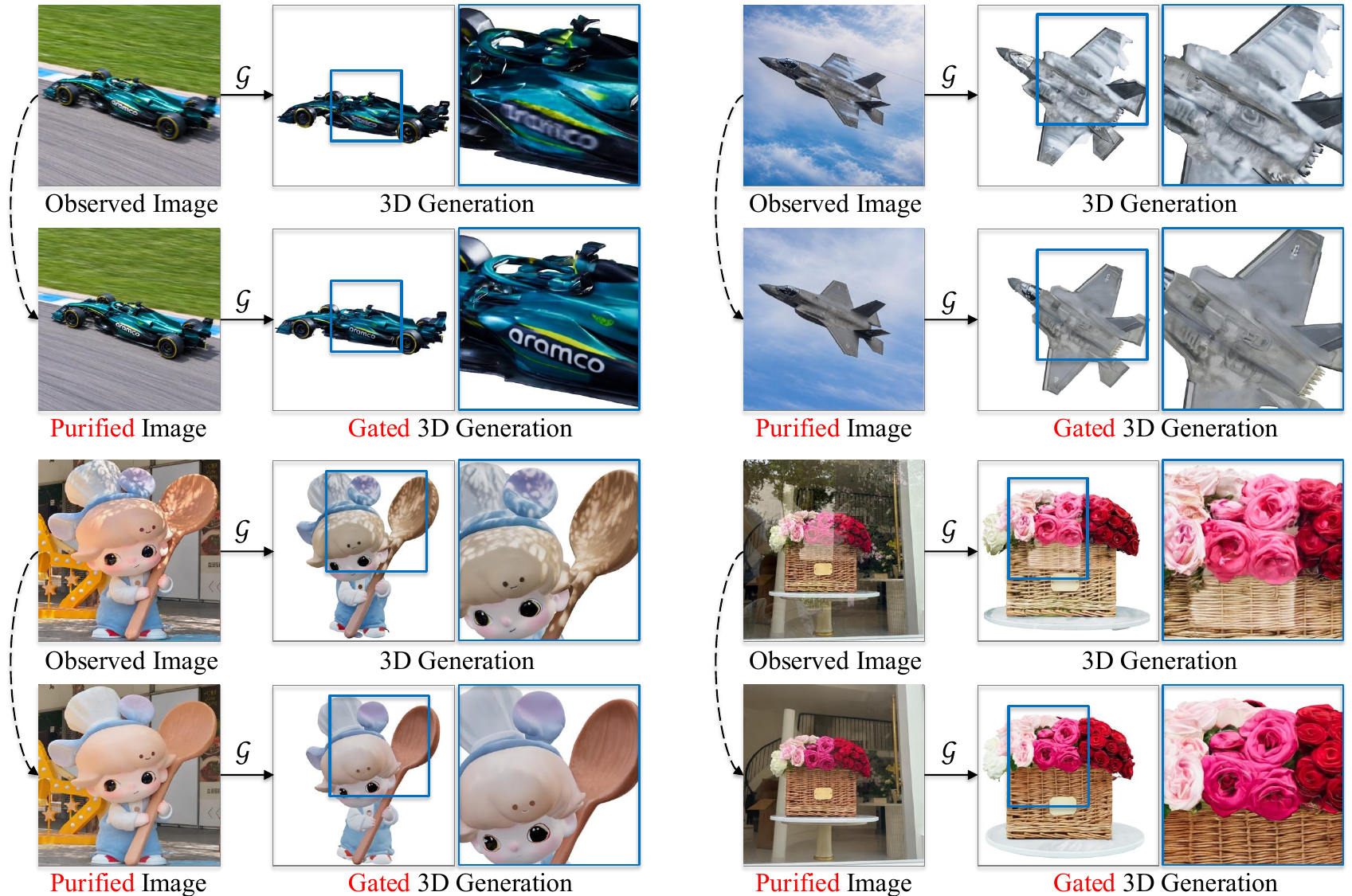}%
}{Gate3DGen improves 3D generation by purifying real-world image input.}{fig:teaser}
\date{}

\author{%
\hypersetup{urlcolor=black}%
\begin{tabular}[t]{@{}l@{\quad}l@{\quad}l@{\quad}l@{}}
\textbf{Xiaolin Li}\textsuperscript{1,2*} &
\textbf{\href{https://inso-13.github.io/}{Zihao Huang}}\textsuperscript{2*} &
\textbf{\href{https://mfp0610.github.io/}{Fanpeng Meng}}\textsuperscript{2} &
\textbf{\href{https://chensjtu.github.io/}{Chen Yang}}\textsuperscript{2}\\[2pt]
\textbf{\href{https://wuguanjun0627.github.io/}{Guanjun Wu}}\textsuperscript{3,2} &
\textbf{\href{https://www.sigs.tsinghua.edu.cn/zk_en/}{Kai Zhang}}\textsuperscript{1}\correspondingmark &
\textbf{\href{https://jaminfong.cn/}{Jiemin Fang}}\textsuperscript{2\ensuremath{\ddagger}\supracorresponding} &
\textbf{\href{https://www.qitian1987.com/index.html}{Qi Tian}}\textsuperscript{2}\correspondingmark
\end{tabular}\\[8pt]
\begin{tabular}[t]{@{}l@{\quad}l@{\quad}l@{}}
\textsuperscript{1}Tsinghua University &
\textsuperscript{2}Huawei Inc. &
\textsuperscript{3}Huazhong University of Science and Technology\\[6pt]
\textsuperscript{*}Equal contribution &
\textsuperscript{\ensuremath{\ddagger}}Project lead &
\correspondingmark Corresponding authors
\end{tabular}\\[7pt]
}

\begin{document}
\maketitle
\clearpage
\flushbottom

\section{Introduction}
\label{sec:introduction}

Single-image 3D generative models have achieved remarkable progress on curated benchmarks~\citep{hong2023lrm,tochilkin2024triposr,xu2024instantmesh,xiang2025trellis2,li2026pixal3d,yang2026supratex}, yet degrade sharply on real-world images.
This vulnerability stems from restrictive assumptions about input images. Contemporary models assume centered, sharp, and fully visible objects under uniform studio lighting~\citep{xu2024instantmesh,xiang2024trellis,xiang2025trellis2}.
In real-world scenarios, however, unconstrained photographs routinely exhibit blur, uneven illumination, or occlusions~\citep{cho2025robust3d,qu2025deocc}, corrupting visual cues and inducing severe artifacts in 3D outputs.

Open-source image-to-3D pipelines commonly use background removal and recentering~\citep{tochilkin2024triposr,xu2024instantmesh,xiang2024trellis,zhao2025hunyuan3d}, which do not directly address photometric corruptions or foreground occlusions.
Specialized preprocessing also targets individual factors such as illumination through delighting~\citep{li2025cue3d,zhao2025hunyuan3d}.
When multiple degradations coexist, correcting one factor can leave other sources of ambiguity in the conditioning image.
Our comparisons show that generic 2D image editing can alter object appearance or contours, with adverse effects on downstream geometry (Section~\ref{sec:three_d_results}).
The objective is therefore to recover a suitable observation of the same object, retaining the visible structural evidence that the 3D generator uses to infer shape and appearance.
This motivates selective input purification that preserves visual cues for 3D generation while avoiding unnecessary changes to already suitable clean inputs.

\looseness=-1
We therefore introduce \methodname{}, a selective pre-processing model that gates real-world images to existing 3D generators (Figure~\ref{fig:teaser}).
It routes each image through purification or an unchanged bypass, keeping the downstream generators frozen.
\methodname{} couples a \emph{degradation-aware router} with a \emph{3D-oriented one-step purifier}.
The router bypasses inputs classified as clean and selects the remaining inputs for purification, while extracting continuous degradation-aware features for the purifier.
Conditioned on these features, the one-step purifier purifies degraded inputs in a single forward pass to reduce inference cost, with spatial losses that encourage preservation of object identity and essential geometric and textural cues.
Training requires matched ground-truth references, which are unavailable for real-world images.
To provide paired training supervision without manual collection, we construct a synthetic data engine from 3D assets.
Rendering multi-view anchors from \NumAssets{} 3D assets~\citep{zhang2025texverse} and synthesizing composite corruptions (\eg, uneven illumination, blur, and occlusions) yields \NumTrainPairs{} aligned image pairs and \NumEvalCases{} held-out cases with 3D ground truth.

We evaluate the purifier on this synthetic benchmark with two pretrained 3D generative models, TRELLIS.2~\citep{xiang2025trellis2} and Pixal3D~\citep{li2026pixal3d}.
For 2D purification, \methodname{} reduces foreground LPIPS by $39.1\%$ while preserving subject silhouettes, with an inference time of $2.3$\,s per image. In comparison, the image editing baselines require $15.6$--$100.8$\,s per image and can alter object identity and reduce reconstruction fidelity (Table~\ref{tab:two_d_quality}).
When used to condition these 3D generative models, the purified images consistently improve all geometry and appearance metrics over the original observations.
Specifically, \methodname{} achieves a $61.5\%$ reduction in all-surface mesh distance, closes $55.5\%$ of the geometry gap to clean references on average (up to $83.1\%$ on TRELLIS.2), and reduces rendering LPIPS by $10.9\%$.
Evaluated separately, the degradation-aware router identifies degraded inputs with $88.45\%$ recall and $93.47\%$ specificity on clean inputs, sending inputs classified as degraded for purification.
On real photographs, VLM scores and human preferences further support identity preservation and degradation correction, complementing the reference-based evaluation on synthetic inputs.

Our contributions are as follows:
\begin{itemize}
    \setlength{\itemsep}{0pt}
    \item We formulate robust single-image 3D generation as selective input purification, improving existing 3D generators through purified inputs without retraining them.
    \item We introduce \methodname{}, combining a router that bypasses inputs classified as clean with a one-step purifier. Degradation-aware features and spatial losses guide purification toward preserving object identity, contours, and geometric and textural cues for downstream 3D generation.
    \item \looseness=-1
We construct a synthetic data engine providing paired purification supervision from 3D assets and a benchmark with 3D ground truth for 3D evaluation. Our method reduces foreground LPIPS by $39.1\%$ and TRELLIS.2 mesh distance by $61.5\%$ relative to degraded inputs, demonstrating gains in image fidelity and downstream 3D geometry.
\end{itemize}

\section{Related Work}
\label{sec:related_work}

\noindent\textbf{Robustness in Single-Image 3D Generation.}
Modern single-image 3D methods, including feed-forward reconstruction and iterative diffusion or flow models, achieve high fidelity on curated inputs~\citep{hong2023lrm,tochilkin2024triposr,xu2024instantmesh,boss2024sf3d,xiang2024trellis,zhao2025hunyuan3d,li2025triposg,team2025hunyuan3d25,xiang2025trellis2,wu2025unilat3d,li2026pixal3d,yang2026supratex}.
Blur, uneven illumination, and occlusions can reduce their generation quality.
Common input preprocessing operations include foreground segmentation and recentering, which do not directly correct illumination, blur, or foreground occlusions.
Specialized methods address illumination through lighting decomposition or delighting~\citep{li2025cue3d,team2025hunyuan3d25}.
Related work investigates recovery of occluded geometry through volumetric reconstruction or multi-view synthesis~\citep{cho2025robust3d,qu2025deocc,shi2025scenemaker}.
Our work focuses on selective input purification before frozen 3D generators, addressing composite degradations involving uneven illumination, blur, and occlusions.

\noindent\textbf{2D Image Editing and Degradation Routing.}
Recent advances in instruction-guided editing~\citep{blackforestlabs2025fluxkontext,wu2025qwenimage,wu2026omnigen2,liu2025step1xedit,zhang2026mageflow,superintelligence2026firered} and all-in-one restoration~\citep{potlapalli2023promptir,conde2024instructir,lin2024diffbir,yu2024supir,guo2024adaptir} synthesize visually compelling images.
Their objectives do not explicitly measure downstream 3D fidelity.
In our comparisons, generic image editing methods can remove visible degradations while changing object appearance or contours, with adverse effects on downstream geometry (Section~\ref{sec:three_d_results}).
Editing clean photographs can introduce unwanted changes, motivating selective restoration. No-reference quality metrics~\citep{mittal2013niqe,yang2022maniqa,zhang2023liqe} and general vision-language models~\citep{you2024depictqawild,bai2025qwen25vl,openai2025gpt5} provide candidate quality signals, but their suitability for routing requires evaluation (Section~\ref{sec:gate_results}).
\methodname couples a degradation-aware router with a one-step purifier to preserve identity and contours while selectively purifying inputs.

\section{Method}
\label{sec:method}

\begin{figure}[!t]
    \centering
    \includegraphics[width=\linewidth,keepaspectratio]{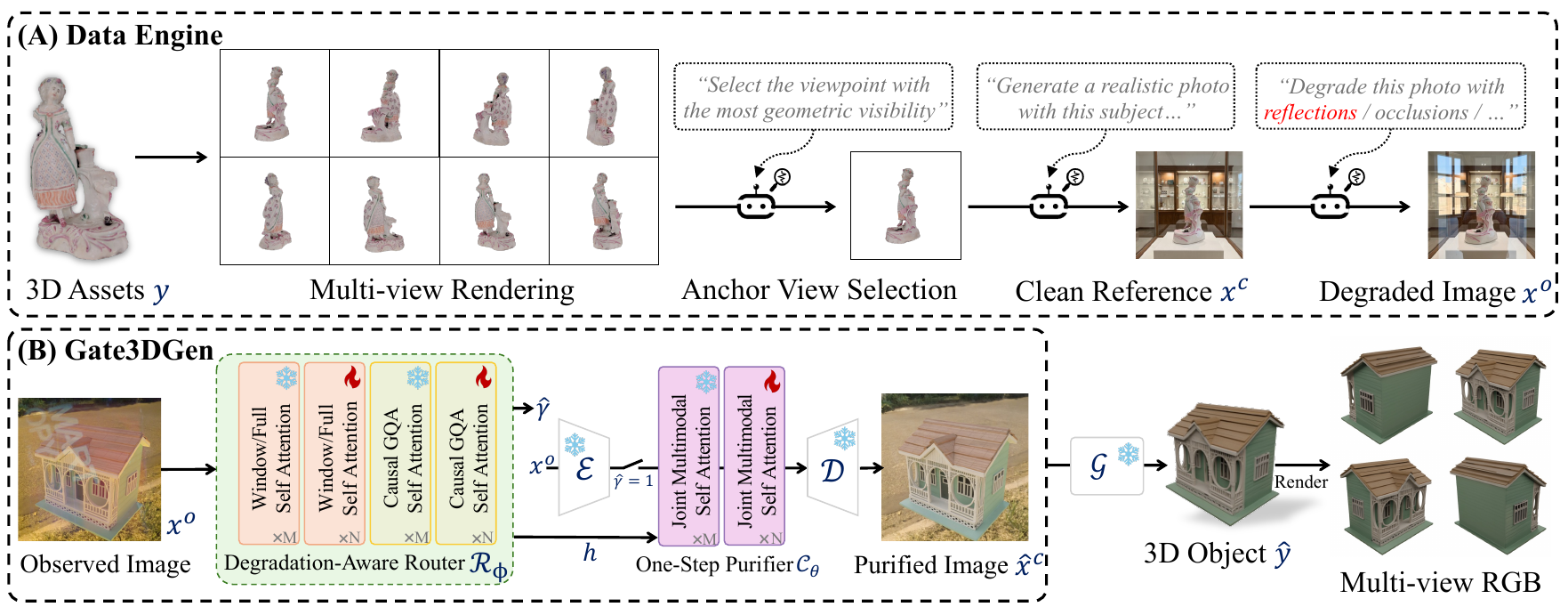}
    \caption{\textbf{Overview of \methodname.}
    \textbf{(A)~Data Engine.} We render 3D assets from multi-view, select informative anchor views, and apply composite degradations to construct aligned observation-reference pairs $(\xo, \xc)$.
    \textbf{(B)~Gate3DGen.} Router $\gate$ bypasses inputs classified as clean ($\widehat{\gamma}=0$). Otherwise ($\widehat{\gamma}=1$), its features $h$ guide the one-step DiT purifier $\canon$. $\xchat$ conditions the frozen 3D generator $\gen$. Clean references and spatial losses are used only in training.}
    \label{fig:protocol_overview}
\end{figure}

\methodname{} is a selective pre-processing model that purifies real-world input images without modifying the downstream 3D generator.
It targets images degraded by uneven illumination, blur, and occlusions while encouraging preservation of geometric and textural cues.
Figure~\ref{fig:protocol_overview} summarizes its data engine, router, and one-step purifier.

\subsection{Problem Formulation}
\label{sec:problem_formulation}

Let $\gen: \mathcal{X} \to \mathcal{Y}$ be a pretrained generator mapping a photograph to a 3D representation $y \in \mathcal{Y}$.
A matched clean reference $\xc$ shows a centered, sharp, fully visible subject under uniform lighting, with $\gen(\xc) \approx y$.
An observed image $\xo$ may be clean or degraded. Uneven illumination, blur, and occlusions can violate these assumptions.
These degradations can distort geometry ($\gen(\xo) \not\approx y$).

We learn to purify input images under supervision from matched clean references.
The purifier maps $\xo$ toward the matched clean reference $\xc$ using reconstruction and spatial losses that encourage preservation of object identity, contours, and textural cues.
The objective is given by
\begin{equation}
\min_{\canon} \; \mathbb{E}_{(\xo, \xc)} \left[ \ell\big(\canon(\xo),\, \xc\big) \right].
\label{eq:editing_objective}
\end{equation}
Here, $\ell$ combines the latent, photometric, perceptual, region-weighted reconstruction, and edge losses defined in Section~\ref{sec:method_backbone}.
To avoid unnecessary purification of clean inputs ($\xo \approx \xc$), the degradation-aware router $\gate$ predicts a routing decision $\widehat{\gamma} \in \{0, 1\}$.
The routing decision determines the pre-processed image as follows
\begin{equation}
    \xchat =
    \begin{cases}
    \xo, & \widehat{\gamma} = 0 \quad (\textsc{Keep}),\\
    \canon(\xo), & \widehat{\gamma} = 1 \quad (\textsc{Purify}),
    \end{cases}.
    \label{eq:framework_interface}
\end{equation}
The \emph{pre-processed image} $\xchat$ is either a \emph{purified image} from $\canon$ (\textsc{Purify}) or the unchanged input (\textsc{Keep}).
To supervise this objective where matched pairs $(\xo, \xc, \gamma)$ are unavailable for real-world images, we construct a synthetic data engine from 3D assets (Section~\ref{sec:data_pipeline}) to supply training data.

\subsection{Synthetic Data Engine}
\label{sec:data_pipeline}

Learning $\canon$ requires paired images $(\xo_i,\xc_i)$ that share identical 3D geometry and viewpoints while differing only in capture quality, whereas learning $\gate$ requires examples labeled by whether purification is necessary.
Because real photographs provide neither matched clean counterparts nor controlled routing labels, we use high-fidelity 3D assets to construct this supervision (Figure~\ref{fig:protocol_overview}A).

\looseness=-1
\noindent\textbf{Anchor view selection.}
From about $4k$ high-fidelity 3D assets in TexVerse~\citep{zhang2025texverse}, we render \NumCanonicalViews{} views under uniform lighting.
A vision-language model (VLM) ranks these views by geometric visibility and self-occlusion.
The \NumViews{} most informative views per asset serve as multi-view anchors and matched clean references $\xc_i$.

\noindent\textbf{Degradation synthesis.}
For each anchor, instruction-guided editing produces a matched observation $\xo_i$ with composite corruptions, including uneven illumination, blur, and occlusions.
Each pair is audited by Qwen3-VL~\citep{qwenteam2025qwen3vl} for structural consistency, and RMBG-2.0 extracts a soft foreground mask $m_i$ from the clean reference $\xc_i$.
This pipeline produces roughly $44k$ image pairs $\mathcal{D}_{\mathrm{train}}=\{(\xo_i,\xc_i,m_i)\}$ for $\canon$ and $100k$ routing pairs $\mathcal{D}_{\mathrm{route}}=\{(x_j,\gamma_j)\}$ for $\gate$, where $\gamma_j=1$ denotes \textsc{Purify} and $\gamma_j=0$ denotes \textsc{Keep}.
Training and evaluation assets are disjoint.

\subsection{Degradation-Aware Router}
\label{sec:repair_gate}

Purifying clean inputs can introduce unwanted changes. To guide purification of images with spatially varying degradations, we use features extracted from the input image.
We therefore design the degradation-aware router $\gate$ based on Qwen2.5-VL~\citep{bai2025qwen25vl}.
Trained on $\mathcal{D}_{\mathrm{route}}$, the router predicts whether purification is needed and extracts continuous degradation-aware features.

\noindent\textbf{Logit-based routing.}
Given input $\xo$, the router returns $(p,h)=\gate(\xo)$, where $p\in[0,1]$ is the predicted probability that purification is needed and $h$ denotes continuous degradation-aware features.
Rather than using greedy decoding, we derive the purification probability $p$ directly from the unnormalized logits for the routing decisions \textsc{Purify} and \textsc{Keep}, denoted by $o_{\mathrm{purify}}$ and $o_{\mathrm{keep}}$.
\begin{equation}
p = \sigma(o_{\mathrm{purify}} - o_{\mathrm{keep}}),
\label{eq:gate_probability}
\end{equation}
where $\sigma(\cdot)$ is the sigmoid function.
We apply a threshold $\tau$ to obtain the routing indicator
\begin{equation}
\widehat{\gamma}=\mathbb{I}[p\ge\tau]\in\{0,1\},
\label{eq:gate_decision_rule}
\end{equation}
where $\mathbb{I}[\cdot]$ is the indicator function.
Predicted clean inputs bypass purification unchanged.

\noindent\textbf{Degradation-aware conditioning.}
When purification is triggered ($\widehat{\gamma}=1$), we extract the final hidden states $h \in \mathbb{R}^{N \times d}$ from $\gate$ under a fixed editing instruction. These features condition $\canon$.

\subsection{3D-Oriented One-Step Purifier}
\label{sec:method_backbone}

The diffusion purifier $\canon$ purifies degraded inputs in a single forward pass with spatial losses that encourage preservation of object identity and geometric and textural cues.
Let $\enc$ and $\dec$ denote the encoder and decoder of its frozen image autoencoder.
We initialize $\canon$ from the released FireRed-Image-Edit-1.1~\citep{superintelligence2026firered} checkpoint, whose architecture and flow-matching formulation follow the FireRed-Image-Edit-1.0 technical report~\citep{superintelligence2026firered}, and condition it on the observation latent $\enc(\xo)$ and the continuous degradation-aware features $h$ produced by the router.

\noindent\textbf{One-step flow-matching.}
Standard flow-matching models~\citep{superintelligence2026firered} define a linear probability path between the target latent $z_0 = \enc(\xc)$ and Gaussian noise $\varepsilon \sim \mathcal{N}(0, I)$, given by
\begin{equation}
z_t = (1 - t)\, z_0 + t\, \varepsilon, \quad t \in [0, 1],
\label{eq:flow_path}
\end{equation}
The velocity network $v_\theta(z_t,\enc(\xo),h,t)$ is trained to regress $\varepsilon-z_0$.
Standard inference integrates from $t = 1$ to $t = 0$ over multiple network evaluations.
We train and evaluate $\canon$ at $t = 1$, where $z_1 = \varepsilon$, to reduce inference latency.
One prediction yields the latent estimate and decoded image
\begin{equation}
\widehat{z}_0 = \varepsilon - v_\theta\big(\varepsilon,\, \enc(\xo),\, h,\, 1\big),
\qquad
\xchat = \dec(\widehat{z}_0),
\label{eq:one_step_endpoint}
\end{equation}
where $\dec$ is the frozen VAE decoder.
Training and inference both use Equation~\ref{eq:one_step_endpoint}, with one evaluation of $v_\theta$ followed by VAE decoding, avoiding iterative denoising during purification.
At this endpoint, the noisy latent contains only Gaussian noise. The observation latent and router features provide the image-specific conditioning. The velocity network is supervised using the clean latent and the decoded image, without iterative sampling.

\noindent\textbf{3D-oriented spatial constraints.}
For a training triple $(\xo_i, \xc_i, m_i) \in \mathcal{D}_{\mathrm{train}}$, let $z_{0,i} = \enc(\xc_i)$, and let $\widehat{z}_{0,i}$ and $\xchat_i$ be obtained from Equation~\ref{eq:one_step_endpoint}.
Standard pixel reconstruction losses treat all regions uniformly.
Generic image editing methods overwrite uncorrupted regions and blur high-frequency shading cues that 3D generators rely on.
The region-weighted reconstruction and edge losses guide purification by supervising the subject, emphasizing degradation-affected regions, and encouraging preservation of edges.
We obtain a soft degradation map $r_i$ by smoothing the pairwise residual between $\xo_i$ and $\xc_i$.
For an error tensor $a$ and a nonnegative spatial weight $w$ broadcast to the shape of $a$, we define the normalized weighted mean
\begin{equation}
\overline{\ell}(a,w)
=
\frac{\sum_j w_{(j)}\lvert a_{(j)}\rvert}
{\max\!\left(\sum_j w_{(j)},1\right)},
\label{eq:normalized_masked_mean}
\end{equation}
where the parenthesized subscript $(j)$ indexes channel and spatial entries within each tensor after broadcasting, while $i$ indexes training samples. With $\delta_i=\xchat_i-\xc_i$, the implemented losses are
\begin{equation}
\mathcal{L}_{\mathrm{region}}=\overline{\ell}(\delta_i,m_i)+\overline{\ell}(\delta_i,r_i),
\qquad
\mathcal{L}_{\mathrm{edge}}=\tfrac12\!\left[\overline{\ell}(D_u\delta_i,m_i^u)+\overline{\ell}(D_v\delta_i,m_i^v)\right],
\label{eq:spatial_losses}
\end{equation}
The two terms in $\mathcal{L}_{\mathrm{region}}$ are normalized independently: the first supervises the full subject through $m_i$, while the second focuses on degradation-affected regions through $r_i$.
The edge term matches finite differences along the horizontal ($u$) and vertical ($v$) image axes, weighted by adjacent subject-mask values.
These weights guide purification, and selectivity refers to image-level routing.
Section~\ref{app:method_details} details degradation-map pooling and adjacent-mask edge weights.

\noindent\textbf{Total training objective.}
Combining spatial constraints with the backbone objectives gives the full purifier loss:
\begin{equation}
\mathcal{L} = \lambda_{\mathrm{lat}} \mathcal{L}_{\mathrm{lat}}
+ \lambda_{1} \mathcal{L}_{1}
+ \lambda_{\mathrm{lpips}} \mathcal{L}_{\mathrm{lpips}}
+ \lambda_{\mathrm{region}} \mathcal{L}_{\mathrm{region}}
+ \lambda_{\mathrm{edge}} \mathcal{L}_{\mathrm{edge}},
\label{eq:training_loss}
\end{equation}
where $\mathcal{L}_{\mathrm{lat}} = \operatorname{MSE}(\widehat{z}_{0,i},z_{0,i})$ supervises latent reconstruction at the endpoint, $\mathcal{L}_{1} = \operatorname{mean}|\xchat_i-\xc_i|$ is the full-image photometric loss, and $\mathcal{L}_{\mathrm{lpips}} = \operatorname{LPIPS}(\xchat_i, \xc_i)$ enforces perceptual agreement.
The latent term supervises the clean representation, while photometric and perceptual terms supervise the decoded image alongside spatial and gradient constraints. Loss weights appear in Section~\ref{app:method_details}.

Clean references and foreground masks are used only during training. Inference uses the observed image and router features. Spatial losses encourage preservation of geometric and textural cues without gradients through the frozen 3D generator, whose output quality is evaluated in Section~\ref{sec:three_d_results}.

\section{Experiments}
\label{sec:experiments}

\providecommand{\dgp}[1]{\ifdim #1pt<0pt \textcolor{red!75!black}{#1\%}\else\textcolor{green!45!black}{+#1\%}\fi}
\providecommand{\dg}[2]{#1\,{\small(\dgp{#2})}}

\subsection{Experimental Setup}
\label{sec:experimental_setup}

\noindent\textbf{Benchmark and evaluation protocols.}
Our benchmark contains \NumEvalCases{} held-out DTC cases~\citep{dong2025digitaltwin} with 3D ground truth and clean references sharing the degraded observations' camera poses.
We compare 3D outputs from degraded, edited, and clean-reference images against ground truth.
Our purifier and all editing baselines process every degraded input. Routing is evaluated separately and does not affect these results.

\noindent\textbf{Baselines.}
We compare against $7$ instruction-guided image editing methods, namely
\underline{FLUX.1-Kontext}~\citep{blackforestlabs2025fluxkontext},
\underline{FireRed-Image-Edit (1.1)}~\citep{superintelligence2026firered},
\underline{Mage-Flow-Edit}~\citep{zhang2026mageflow},
\underline{OmniGen2}~\citep{wu2026omnigen2},
\underline{Qwen-Image-Edit}~\citep{wu2025qwenimage},
\underline{Step1X-Edit}~\citep{liu2025step1xedit},
and \underline{GPT Image-2.5 (Flare)}~\citep{openai2026gptimage2}.
We evaluate downstream 3D quality with Pixal3D~\citep{li2026pixal3d} and TRELLIS.2~\citep{xiang2025trellis2}.

\noindent\textbf{Evaluation metrics.}
For \emph{degradation-aware routing}, we report \textsc{Purify}-class recall, overall accuracy, and $\mathrm{H}$, defined as their harmonic mean, all expressed as percentages (Table~\ref{tab:gate_results}).
For \emph{2D image fidelity}, we report foreground LPIPS, PSNR, SSIM, Sobel gradient MAE, and foreground mask IoU against the clean reference.
For \emph{3D metrics}, we report all-surface Mesh Distance (MD)~\citep{xiang2025trellis2} and $F_1^{\mathrm{all}}$ ($F_1@10^{-8}$), outer-surface Chamfer Distance (CD)~\citep{fan2017pointset} and $F_1^{\mathrm{outer}}$ ($F_1@10^{-6}$) after ICP alignment to the ground-truth asset, PSNR-N~\citep{wang2004ssim} on rendered normal maps, and LPIPS~\citep{zhang2018lpips} on four-view shaded and albedo renderings.

\begin{table}[!t]
    \caption{\textbf{Degradation-aware routing on synthetic and real images (\%).}
    Recall is for \textsc{Purify}. Accuracy is overall (\textsc{Purify} and \textsc{Keep}). $\mathrm{H}=2\,\mathrm{Recall}\ \times \mathrm{Accuracy}/(\mathrm{Recall}+\mathrm{Accuracy})$.
    }
\label{tab:gate_results}
\centering
\small
\renewcommand{\arraystretch}{1.15}
\setlength{\tabcolsep}{14pt}
\begin{suprafulltable}{l|ccc|ccc}{12}
\toprule
& \multicolumn{3}{c}{\textbf{Benchmark}} & \multicolumn{3}{c}{\textbf{In-the-wild}}\\
\cmidrule(lr){2-4}\cmidrule(lr){5-7}
Method & Recall & Accuracy & H$\uparrow$ & Recall & Accuracy & H$\uparrow$\\
\midrule
mPLUG-Owl2 & 100.00 & 54.12 & 70.23 & 0.00 & 41.67 & 0.00\\
DepictQA-Wild & 99.54 & 53.42 & 69.53 & 100.00 & 58.33 & \textbf{73.68}\\
LIQE-MIX & 51.66 & 65.48 & 57.75 & 17.86 & 48.96 & 26.17\\
NIQE & 98.52 & 53.32 & 69.19 & 75.00 & 54.17 & 62.91\\
Qwen2.5-VL & 83.64 & 74.21 & 78.64 & 14.29 & 43.75 & 21.54\\
MANIQA & 79.57 & 89.04 & 84.04 & 12.50 & 45.83 & 19.64\\
GPT-5 & 72.27 & 77.58 & 74.83 & -- & -- & --\\
\midrule
\rowcolor{supraTableWash}
\textbf{Ours} & 88.45 & 90.77 & \textbf{89.59} & 89.29 & 58.33 & \underline{70.56}\\
\bottomrule
\end{suprafulltable}

\end{table}

\subsection{Degradation-Aware Routing Results}
\label{sec:gate_results}

Table~\ref{tab:gate_results} reports routing results on $\NumGateDegraded{}$ degraded and $\NumGateClean{}$ clean synthetic inputs and on $\NumWildCases{}$ in-the-wild images. Because real photographs lack explicit routing labels, a GPT-5 agent assigns proxy \textsc{Purify}/\textsc{Keep} labels by comparing multi-view renderings and no-reference metrics for 3D assets generated from the unedited and purified images by Pixal3D and TRELLIS.2.
The router should direct degraded inputs to the purifier and leave clean inputs unchanged, as uncorrected degradations can introduce geometric artifacts (Table~\ref{tab:three_d_quality}).
\textbf{On the synthetic benchmark}, the router attains $88.45\%$ recall and $90.77\%$ accuracy, yielding the highest $\mathrm{H}$ of $89.59\%$, which exceeds the strongest baseline, MANIQA~\citep{yang2022maniqa}, by $5.55$ percentage points.
NIQE~\citep{mittal2013niqe}, mPLUG-Owl2~\citep{ye2024mplugowl2}, and DepictQA-Wild~\citep{you2024depictqawild} exceed $98\%$ recall, but their accuracies remain within one percentage point of the $53.7\%$ obtained by outputting \textsc{Purify} for all inputs, resulting in $\mathrm{H}$ values of only $69.19$--$70.23\%$.
MANIQA attains $89.04\%$ accuracy but only $79.57\%$ recall. Our router improves both, indicating that its higher $\mathrm{H}$ reflects better detection of inputs needing purification alongside more accurate routing overall.
Zero-shot Qwen2.5-VL, which shares the backbone of our router, reaches $\mathrm{H}=78.64\%$. Training on $\mathcal{D}_{\mathrm{route}}$ raises recall by $4.81$ percentage points and accuracy by $16.56$ points, improving $\mathrm{H}$ by $10.95$ points.
\textbf{On the in-the-wild benchmark}, the router ranks second in $\mathrm{H}$ ($70.56\%$), with $89.29\%$ recall and $58.33\%$ accuracy against the proxy labels.
DepictQA-Wild achieves the highest $\mathrm{H}$ ($73.68\%$), but its $100\%$ recall and $58.33\%$ accuracy match the strategy of always predicting \textsc{Purify}.
Thus, a high $\mathrm{H}$ alone does not establish effective preservation of \textsc{Keep} inputs, and our router does not outperform this trivial strategy under the real-image proxy protocol.
The high recall and lower accuracy are consistent with over-routing of inputs labeled \textsc{Keep} to purification. This gap may reflect the use of clean renders for the \textsc{Keep} class in $\mathcal{D}_{\mathrm{route}}$. Whether adding real photographs to this class improves routing remains to be tested.

\nocite{openai2025gpt5}

\begin{table}[!t]
\caption{\textbf{Synthetic 2D purification.}
Fidelity metrics use clean references. Input is unedited, Grad.\ is Sobel gradient MAE, and IoU is foreground mask overlap.
 $^{*}$ denotes the commercial closed-source model.}
\label{tab:two_d_quality}
\centering
\small
\renewcommand{\arraystretch}{1.15}
\setlength{\tabcolsep}{20pt}
\begin{suprafulltable}{l|c|ccccc}{10}
\toprule
Method & Time (s)$\downarrow$ & LPIPS$\downarrow$ & PSNR$\uparrow$ & SSIM$\uparrow$ & Grad.$\downarrow$ & IoU$\uparrow$\\
\midrule
Degraded (Input) & -- & 0.2262 & 20.254 & 0.8323 & 0.0584 & 0.9537\\
\midrule
FLUX.1 & 30.2 & 0.2577 & 17.992 & \underline{0.7907} & \underline{0.0688} & 0.9373\\
OmniGen2 & 84.9 & 0.4166 & 13.491 & 0.6478 & 0.1123 & 0.6238\\
Step1X & 100.8 & 0.2426 & 18.828 & 0.7761 & 0.0781 & 0.8971\\
Qwen-Edit & 81.5 & 0.2389 & 18.710 & 0.7812 & 0.0727 & 0.9537\\
Mage-Flow & \underline{15.6} & 0.2418 & \underline{19.175} & 0.7755 & 0.0835 & 0.9530\\
GPT Image-2.5 Flare$^{*}$ & -- & 0.2710 & 16.571 & 0.7020 & 0.1319 & 0.9189\\
FireRed & 83.2 & \underline{0.2258} & 15.868 & 0.7677 & 0.0814 & \underline{0.9604}\\
\midrule
\rowcolor{supraTableWash}
\textbf{Ours} & \textbf{2.3} & \textbf{0.1378} & \textbf{22.514} & \textbf{0.8708} & \textbf{0.0488} & \textbf{0.9758}\\
\bottomrule
\end{suprafulltable}

\end{table}

\subsection{2D Purification Results}
\label{sec:two_d_results}

Table~\ref{tab:two_d_quality} reports fidelity to clean references. All seven generic editors lower PSNR and SSIM and raise gradient MAE by $17.8\%$--$125.9\%$ relative to unedited inputs. Only FireRed~\citep{superintelligence2026firered} lowers LPIPS, by $0.0004$. Degradation removal thus comes with reduced fidelity to clean references. OmniGen2~\citep{wu2026omnigen2} also reduces silhouette IoU.
\methodname{} alone improves all five metrics, lowering LPIPS by $39.1\%$ and gradient MAE by $16.4\%$ while raising IoU to $0.9758$. The gains over the FireRed backbone support the benefit of our purification training (Section~\ref{sec:method_backbone}).

\noindent\textbf{Inference efficiency.}
\methodname{} achieves the best fidelity across all five metrics with an inference time of $2.3$\,s, compared with $15.6$--$100.8$\,s for the $6$ baselines. This corresponds to a $6.8\times$ speedup over Mage-Flow, the fastest timed baseline, and a $36.2\times$ speedup over FireRed. Our one-step purifier achieves higher image fidelity and lower inference time than the compared image editing methods.

\begin{table}[!t]
    \caption{\textbf{Downstream 3D metrics on TRELLIS.2.}
    Input is degraded, and Oracle uses clean references.
    MD, CD, and $F_1^{\mathrm{all}}$ are $\times10^{-3}$, and $F_1^{\mathrm{outer}}$ is $\times10^{-1}$.
    PSNR-N evaluates normals. LPIPS compares four-view shaded and albedo renderings.
     $^{*}$ denotes a commercial closed-source model.}
    \label{tab:three_d_quality}
    \label{tab:pbr_quality}
    \centering
    \small
\renewcommand{\arraystretch}{1.15}
    \setlength{\tabcolsep}{6.9pt}
    \begin{suprafulltable}{l|ccccc|cc}{14}
    \toprule
    Method & MD$\downarrow$ & $F_1^{\mathrm{all}}$$\uparrow$ & CD$\downarrow$ & $F_1^{\mathrm{outer}}$$\uparrow$ & PSNR-N$\uparrow$ & LPIPS$\downarrow$ & Albedo LPIPS$\downarrow$\\
    \midrule
    Clean reference (Oracle) & 0.517 & 12.46 & 0.517 & 1.1586 & 16.36 & 0.1548 & 0.1575\\
    Degraded (Input) & 1.984 & 11.52 & 1.985 & 1.0531 & 15.26 & 0.1857 & 0.1916\\
    \midrule
    FLUX.1 & 1.528 & 11.06 & 1.528 & 1.0238 & 15.35 & 0.1864 & 0.1935\\
    OmniGen2 & 3.485 & 6.88 & 3.485 & 0.6407 & 13.11 & 0.2362 & 0.2411\\
    Step1X & 1.228 & 10.48 & 1.229 & 0.9637 & 15.25 & 0.1803 & 0.1861\\
    Qwen-Image-Edit & 1.047 & 10.81 & 1.047 & 1.0123 & 15.69 & 0.1780 & 0.1859\\
    Mage-Flow & 1.386 & \underline{11.42} & 1.386 & \underline{1.0422} & 15.39 & 0.1797 & 0.1855\\
    GPT Image-2.5 Flare$^{*}$ & 1.179 & 10.39 & 1.179 & 0.9720 & 15.29 & 0.1792 & 0.1859\\
    FireRed & \underline{0.835} & 10.58 & \underline{0.835} & 0.9889 & \underline{15.82} & \underline{0.1701} & \underline{0.1781}\\
    \midrule
    \rowcolor{supraTableWash}
    \textbf{Ours} & \textbf{0.764} & \textbf{11.96} & \textbf{0.765} & \textbf{1.1205} & \textbf{15.98} & \textbf{0.1654} & \textbf{0.1714}\\
    \bottomrule
    \end{suprafulltable}

\end{table}

\subsection{Downstream 3D Generation Results}
\label{sec:three_d_results}
\label{sec:pbr_results}

\begin{figure}[!t]
\centering
\includegraphics[width=1.0\linewidth,keepaspectratio]{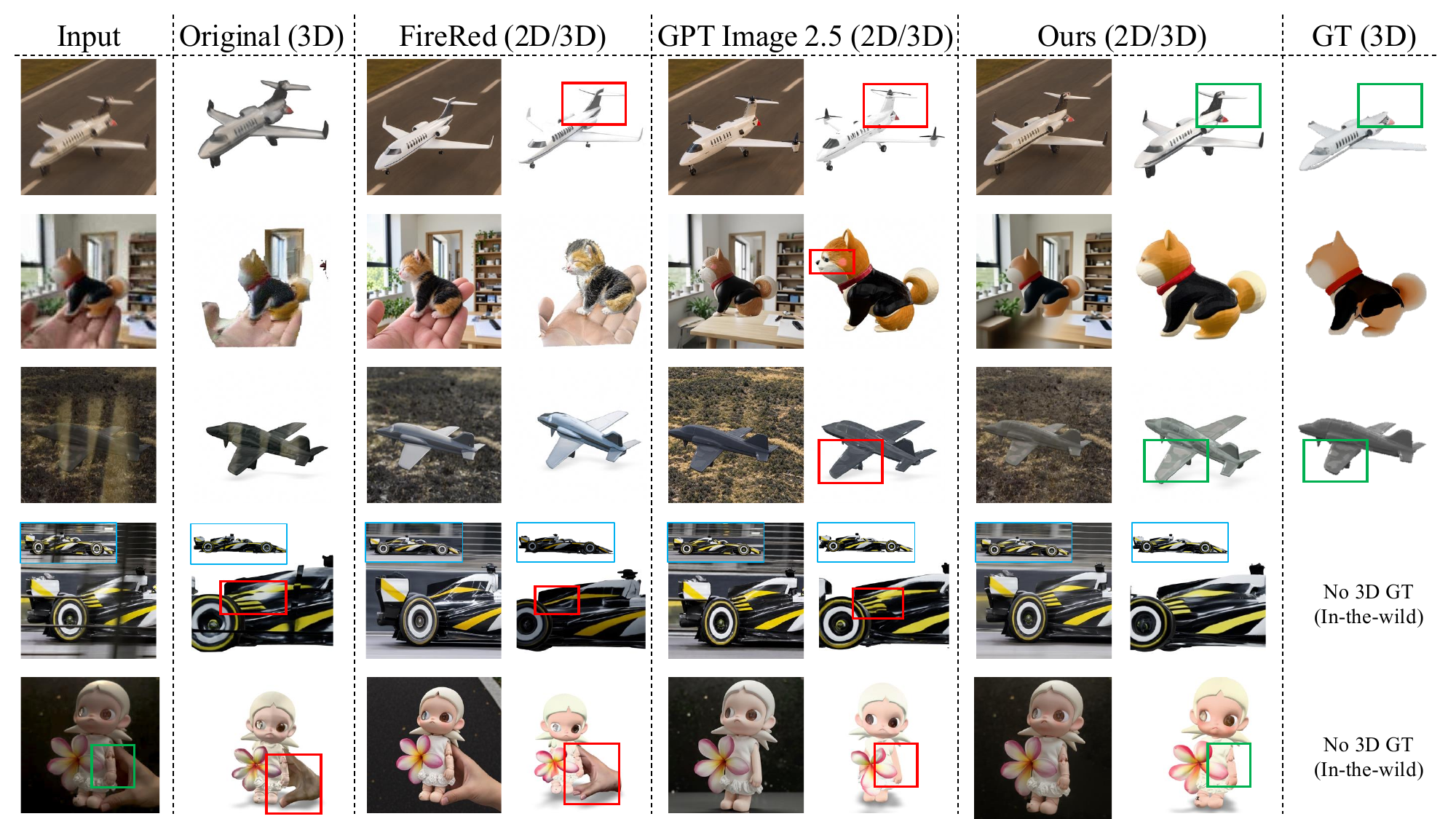}
\caption{\textbf{Comparison of images (2D) and Pixal3D outputs (3D).}
Original (3D) uses unedited inputs. Ours (2D) uses purified images. Boxes mark structure and appearance changes.}
\label{fig:qual_baseline}
\end{figure}

On TRELLIS.2~\citep{xiang2025trellis2} (Table~\ref{tab:three_d_quality}), \methodname{} lowers MD and CD by $61.5\%$ relative to the degraded input, raises $F_1^{\mathrm{all}}$ and $F_1^{\mathrm{outer}}$ by $3.8\%$ and $6.4\%$, and lowers shaded and albedo LPIPS by $10.9\%$ and $10.6\%$, closing $83.1\%$ of the MD gap to the clean-reference oracle.
The baselines reveal a trade-off, as six of the seven image editing methods reduce MD yet all seven lower both $F_1$ scores below the unedited input.
Large geometric artifacts, such as an occluder fused into the object (Figure~\ref{fig:qual_baseline}), can increase MD, whereas $F_1$ at a small distance threshold is sensitive to local geometric deviations.
Generic image editing can therefore reduce large artifacts while introducing local geometric deviations. Only \methodname{} improves MD and both $F_1$ scores among the compared editors, consistent with the spatial constraints in Equation~\ref{eq:spatial_losses}.
Albedo LPIPS also improves, consistent with removing input illumination effects before they are incorporated into the generated material, while the residual gap to the oracle is largest for $F_1^{\mathrm{all}}$ ($11.96$ against $12.46$), as geometry hidden by occlusion is only partially recoverable from a single view.

Figure~\ref{fig:qual_baseline} shows Pixal3D reconstructions. Unedited inputs fuse the hand into the doll. FireRed retains the occluder or recolors the teapot, while GPT Image-2.5 Flare removes degradations but re-synthesizes objects (red boxes). \methodname{} purifies inputs and retains the objects (green boxes).

\begin{table}[!t]
\caption{\textbf{In-the-wild comparisons against 2D editing baselines.}
FD-DINOv2 is calculated on multi-view renders per method. VLM scores span $0$--$10$. Purification measures degradation removal in both VLM evaluation and the user study. Pref\% reports preferences for \methodname{} over each baseline. $^{*}$ marks commercial closed-source models.}
\label{tab:wild_quality}
\label{tab:wild_vlm}
\centering
\small
\renewcommand{\arraystretch}{1.15}
\setlength{\tabcolsep}{1.5pt}
\begin{suprafulltable}{l|cc|ccc|ccc}{16}
\toprule
& \multicolumn{2}{c}{FD-DINOv2$\downarrow$}
& \multicolumn{3}{c}{VLM Scores$\uparrow$}
& \multicolumn{3}{c}{User Study (Pref\%)$\uparrow$}\\
\cmidrule(lr){2-3}\cmidrule(lr){4-6}\cmidrule(lr){7-9}
Method & Pixal3D & TRELLIS.2 & Identity & Purification & Quality & Identity & Purification & Quality\\
\midrule
FLUX.1 & 0.795 & 0.790 & 7.31 & 6.22 & 6.24 & 77.5\% & 81.0\% & 72.1\%\\
OmniGen2 & 0.893 & 0.902 & 5.43 & 4.51 & 4.72 & 86.0\% & 93.2\% & 92.3\%\\
Step1X & 0.788 & 0.789 & 7.91 & 7.09 & 5.98 & 81.8\% & 83.3\% & 86.8\%\\
Qwen-Image-Edit & 0.792 & 0.810 & 8.09 & 6.99 & \underline{6.60} & 91.7\% & 94.6\% & 82.1\%\\
Mage-Flow & 0.739 & 0.769 & \underline{8.50} & \underline{7.55} & 6.22 & 84.2\% & 88.1\% & 80.0\%\\
\shortstack[l]{GPT Image-2.5 Flare$^{*}$} & \underline{0.733} & \textbf{0.722} & 6.95 & 6.31 & \textbf{6.92} & 58.5\% & 57.1\% & 57.5\%\\
FireRed & 0.771 & 0.796 & 8.00 & 7.08 & \textbf{6.92} & 66.7\% & 63.6\% & 63.6\%\\
\midrule
\rowcolor{supraTableWash}
\textbf{Ours} & \textbf{0.716} & \underline{0.757} & \textbf{8.62} & \textbf{8.00} & 6.58 & -- & -- & --\\
\bottomrule
\end{suprafulltable}
\end{table}

\subsection{In-the-Wild Results}
\label{sec:wildcase_results}

We evaluate $\NumWildCases{}$ mobile and web photographs without ground-truth meshes using VLM assessment, human preferences, and FD-DINOv2~\citep{stein2023dinoeval}. Qwen3.8-27B serves as the VLM judge~\citep{chen2024mllmjudge}, scoring Identity (object preservation), Purification (degradation removal), and Quality (asset quality independent of the input). Identity and Purification average 2D and 3D scores, while Quality averages 3D scores over four views and both frozen generators.

For FD-DINOv2, we fit Gaussians per generator to L2-normalized DINOv2 ViT-L/14-reg CLS features from the $\NumWildCases{}$ photographs and $384$ shaded renders (four per asset). Images are resized to $224\times224$ and ImageNet-normalized. Rendered views are separate samples without pooling.

In Table~\ref{tab:wild_quality}, \methodname{} leads in VLM Identity ($8.62$) and Purification ($8.00$). FireRed and GPT Image-2.5 Flare score higher on Quality ($6.92$ versus $6.58$), but lower on Identity ($8.00$ and $6.95$, respectively). The VLM scores support object preservation and degradation removal.

\looseness=-1
The human study uses $341$ valid ratings from pairwise comparisons with each image editing baseline. Preferences for \methodname{} exceed $50\%$ on all three criteria against every baseline. Rates are lowest against GPT Image-2.5 Flare ($58.5\%$, $57.1\%$, and $57.5\%$ for Identity, Purification, and Quality). Human Quality preferences differ from the VLM ranking.

\looseness=-1
Among image editing methods, \methodname{} ranks first in FD-DINOv2 on Pixal3D and second on TRELLIS.2, but none improves over unedited inputs. FD-DINOv2 compares feature distributions of inputs and renders, without directly measuring degradation removal or per-object fidelity. Interpret its rankings alongside VLM scores and human preferences.

\begin{table}[!t]
    \caption{\textbf{Purifier ablations on our synthetic benchmark.}
    We report 2D image fidelity and downstream 3D metrics from TRELLIS.2.
    Checkmarks indicate the use of router features ($h$) and spatial losses (region-weighted reconstruction and edge).
    MD, CD, and $F_1^{\mathrm{all}}$ are $\times10^{-3}$, and $F_1^{\mathrm{outer}}$ is $\times10^{-1}$.
    Shaded and albedo LPIPS use the corresponding four-view renderings.
    }
    \label{tab:ablate_editor}
    \centering
    \small
\renewcommand{\arraystretch}{1.15}
    \setlength{\tabcolsep}{6pt}
    \begin{suprafulltable}{c|cc|cc|cccccc}{20}
    \toprule
    Method & $h$ & \shortstack{Spatial\\losses} & \shortstack{2D\\LPIPS$\downarrow$} & \shortstack{2D\\PSNR$\uparrow$} & MD$\downarrow$ & $F_1^{\mathrm{all}}$$\uparrow$ & CD$\downarrow$ & $F_1^{\mathrm{outer}}$$\uparrow$ & \shortstack{Shaded\\LPIPS$\downarrow$} & \shortstack{Albedo\\LPIPS$\downarrow$}\\
    \midrule
    A & & & \textbf{0.1351} & 22.292 & 0.820 & \textbf{12.06} & 0.820 & 1.1012 & 0.1794 & 0.1781\\
    B & $\checkmark$ & & \underline{0.1363} & \textbf{22.572} & \textbf{0.755} & \underline{11.96} & \textbf{0.755} & \textbf{1.1210} & \underline{0.1658} & \underline{0.1716}\\
    \rowcolor{supraTableWash}
    Full & $\checkmark$ & $\checkmark$ & 0.1378 & \underline{22.514} & \underline{0.764} & \underline{11.96} & \underline{0.765} & \underline{1.1205} & \textbf{0.1654} & \textbf{0.1714}\\
    \bottomrule
    \end{suprafulltable}

\end{table}

\subsection{Ablation Studies}
\label{sec:ablations}

\looseness=-1
\noindent\textbf{3D-oriented one-step purifier.}
Table~\ref{tab:ablate_editor} compares router features $h$ and spatial losses, with $\mathcal{L}_{1}$ retained in all variants.
A removes both components, B retains only $h$, and Full uses both.
A achieves the best 2D LPIPS ($0.1351$) and $F_1^{\mathrm{all}}$ ($12.06$), but the worst MD, CD, $F_1^{\mathrm{outer}}$, and rendering LPIPS scores.
Its rendered LPIPS ($0.1794$) is comparable to Mage-Flow and GPT Image-2.5 Flare (Table~\ref{tab:three_d_quality}), despite at least $0.1$ lower 2D LPIPS. Thus, image similarity alone is insufficient to predict 3D quality.
Adding $h$ lowers MD from $0.820$ to $0.755$ ($7.9\%$) and rendered LPIPS from $0.1794$ to $0.1658$ ($7.6\%$), while increasing 2D LPIPS by $0.0012$.
With the spatial losses, Full has slightly higher MD ($0.764$ versus $0.755$) and lower 2D PSNR than Variant B, but achieves the lowest shaded and albedo LPIPS ($0.1654$ and $0.1714$).
Full also preserves contours more faithfully than B, supporting this configuration. Sections~\ref{app:router_ablation} and~\ref{app:pixal3d} report routing and Pixal3D results.

\FloatBarrier
\section{Conclusion}
\label{sec:conclusion}

Trained on image pairs synthesized from 3D assets, \methodname{} combines a degradation-aware router and a 3D-oriented one-step purifier before frozen 3D generators.
On \NumEvalCases{} cases, its purifier outperforms seven image editing methods in image fidelity and surface geometry for TRELLIS.2 and Pixal3D, closing up to $83.1\%$ of the mesh-distance gap to clean-reference inputs.
VLM scores and human preferences on real photographs support identity preservation and degradation correction.
Ablations and real-image comparisons motivate evaluating purification through downstream surface geometry, rendered appearance, and object fidelity beyond image similarity or visual quality alone.

\pagebreak[4]
\raggedbottom
\setlength{\bibsep}{3pt plus 1pt minus 1pt}
\bibliography{main}
\bibliographystyle{supra3d}

\clearpage
\appendix
\begin{center}
  {\Large\bfseries Gate3DGen: A Pre-processing Model that Selectively Purifies Real-World Images for 3D Generation \par}
  \vspace{0.5em}
  {\large\bfseries Supplementary Materials\par}
\end{center}
\suppressfloats[t]

\section{Method Details}
\label{app:method_details}

Under a fixed prompt, the degradation-aware router is trained to output the \textsc{Purify}/\textsc{Keep} decision described in Section~\ref{sec:repair_gate}. These decisions correspond to the implementation tokens \texttt{Purify} and \texttt{Keep}, respectively.
It updates language-transformer layers 24--27 and the final language normalization. The language-model head and visual backbone are frozen except for rank-16 LoRA adapters in vision blocks 28--31. Training uses token-level supervision with higher weights on the decision tokens and the end-of-sequence token, and with higher sample weights on rare degradation families and on subtle degradations. The discrete routing indicator $\widehat{\gamma}_i$ follows Equation~\ref{eq:gate_decision_rule}, with the threshold $\tau$ frozen on the development set. The language-only router variant omits the vision LoRA adapters.

For the purifier $\canon$, image-transformer blocks 1--28 remain frozen and blocks 29--49 are updated together with the cross-attention projection layers receiving $h$. Training uses the flow-matching endpoint at $t=1$ in Equation~\ref{eq:flow_path}, estimates $\widehat{z}_0$, and reconstructs $\xchat=\dec(\widehat z_0)$ by Equation~\ref{eq:one_step_endpoint}. The subject mask $m_i$ is the 8-bit soft alpha predicted by RMBG-2.0 from $\xc_i$. It is normalized to $[0,1]$ and receives the same resize and random crop as the paired images. The degradation map $r_i$ averages $\lvert\xo_i-\xc_i\rvert/2$ over RGB channels and applies $7\times7$ average pooling with stride 1 and padding 3. In Equation~\ref{eq:normalized_masked_mean}, the single-channel weight is broadcast over RGB channels, and each weighted sum is divided by its own broadcast weight sum, clamped below at 1. For the edge term, $D_u$ and $D_v$ are one-pixel forward differences, with $m_i^u$ and $m_i^v$ given by the products of the two subject-mask values adjacent to each horizontal or vertical edge, respectively.
The loss weights in Equation~\ref{eq:training_loss} are $\lambda_{\mathrm{lat}}=0.1$, $\lambda_{1}=1$, $\lambda_{\mathrm{lpips}}=0.5$, $\lambda_{\mathrm{region}}=1$, and $\lambda_{\mathrm{edge}}=1$.
We train at $1024\times1024$ resolution for 15,000 updates on 32 GPUs (global batch size 32, BF16 precision, AdamW optimizer, cosine learning-rate schedule, peak learning rate $5\times10^{-7}$). At inference, image purification uses a single forward pass with guidance scale 1.0, an empty negative prompt, and a fixed random seed.

\section{Router Ablation}
\label{app:router_ablation}

Table~\ref{tab:ablate_gate} compares the full router with an MLP on VAE latents and with a language-only variant that omits vision LoRA.
The MLP achieves $53.42\%$ recall.
Removing vision LoRA raises recall to $97.04\%$ but lowers accuracy to $75.30\%$, indicating a stronger bias toward \textsc{Purify}.
The full router achieves $88.45\%$ recall and $90.77\%$ accuracy. Vision LoRA reduces this bias with high recall.

\begin{table}[!hb]
    \caption{\textbf{Router ablations on synthetic evaluation inputs.}
    Variants use an MLP on VAE latents or omit vision LoRA. Recall and accuracy are percentages. \textbf{Bold} marks column maxima.}
    \label{tab:ablate_gate}
    \centering
    \small
\renewcommand{\arraystretch}{1.15}
    \setlength{\tabcolsep}{8pt}
    \clipbox{\tabcolsep{} 0pt \tabcolsep{} 0pt}{\begin{tabular}{lcc}
    \toprule
    Method & Recall$\uparrow$ & Accuracy$\uparrow$\\
    \midrule
    MLP on VAE latents & 53.42 & 71.48\\
    Without vision LoRA & \textbf{97.04} & 75.30\\
    \rowcolor{supraTableWash}
    Full router & 88.45 & \textbf{90.77}\\
    \bottomrule
    \end{tabular}}
\end{table}

\section{Relative Performance across 2D and 3D Metrics}
\label{app:radar_comparison}

\begin{figure}[!t]
    \centering
    \includegraphics[width=0.6\textwidth,keepaspectratio]{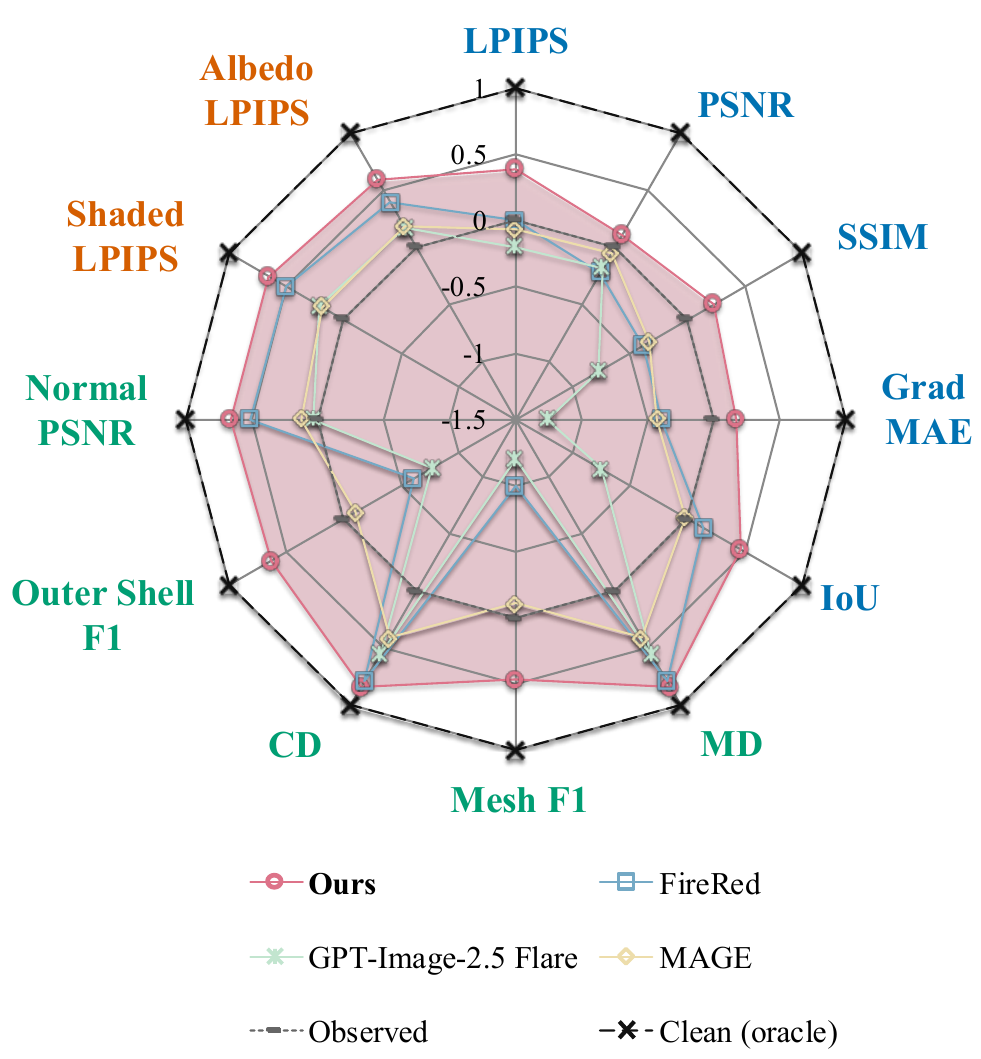}
    \vspace{20pt}
    \caption{\textbf{Oracle-normalized comparison of 2D purification and downstream 3D generation.}
    The radar plot summarizes percentage progress from the observed degraded input (0) toward the matched clean-reference oracle (1) using the metrics reported in Tables~\ref{tab:two_d_quality} and~\ref{tab:three_d_quality}. Negative values indicate deterioration relative to the observed input.
    Blue labels denote 2D image metrics, green labels denote 3D geometry and normal metrics, and orange labels denote rendered-appearance metrics.
    Lower-is-better metrics are direction-adjusted so that farther outward is always better.}
    \label{fig:radar_comparison}
\end{figure}

Figure~\ref{fig:radar_comparison} provides a unified view of the 2D purification metrics in Table~\ref{tab:two_d_quality} and the downstream TRELLIS.2 metrics in Table~\ref{tab:three_d_quality}.
The observed degraded input is the zero reference and the matched clean input is the oracle reference at one. The direction of lower-is-better metrics is reversed so that larger radial values consistently indicate better performance.
Among the representative editing baselines shown, \methodname{} makes positive progress toward the clean oracle on every metric.
In contrast, FireRed, GPT Image-2.5 Flare, and Mage-Flow exhibit negative changes on several 2D fidelity or fine-grained 3D geometry measures, illustrating the trade-off between visible degradation removal and preservation of reconstruction cues.

\section{Additional Pixal3D Results}
\label{app:pixal3d}

The main paper reports downstream 3D metrics on TRELLIS.2 (Table~\ref{tab:three_d_quality}).
Table~\ref{tab:pixal3d_quality} repeats the same evaluation on Pixal3D~\citep{li2026pixal3d}, using the same $\NumEvalCases{}$ cases and the same metric definitions.
\methodname{} obtains the best MD, $F_1^{\mathrm{all}}$, CD, $F_1^{\mathrm{outer}}$, and albedo LPIPS among image editing methods.
Despite these gains in surface geometry, \methodname{} does not achieve the best results on all rendering metrics on Pixal3D.
FireRed obtains higher normal-map PSNR and lower shaded LPIPS, while our albedo LPIPS is only marginally lower ($0.2773$ versus $0.2775$).
Despite gains in surface geometry, improvements in rendered appearance are smaller on Pixal3D than on TRELLIS.2.

\begin{table}[!t]
    \caption{\textbf{Downstream 3D metrics on Pixal3D.}
    Input and Oracle use degraded and clean-reference images, respectively. MD and CD are $\times10^{-3}$.
    $F_1^{\mathrm{all}}$ is all-surface $F_1@10^{-8}$ ($\times10^{-3}$).
    $F_1^{\mathrm{outer}}$ is outer-surface $F_1@10^{-6}$ ($\times10^{-1}$).
    PSNR-N evaluates rendered normals.
    LPIPS and albedo LPIPS use four-view shaded and albedo renderings, respectively.
    \textbf{Bold} and \underline{underline} mark the best and second-best editing results, excluding Input and Oracle. $^{*}$ denotes a commercial closed-source model.}
    \label{tab:pixal3d_quality}
    \begin{center}
    \small
\renewcommand{\arraystretch}{1.15}
    \setlength{\tabcolsep}{5.5pt}
    \begin{suprafulltable}{l|ccccc|cc}{14}
    \toprule
    Method & MD$\downarrow$ & $F_1^{\mathrm{all}}$$\uparrow$ & CD$\downarrow$ & $F_1^{\mathrm{outer}}$$\uparrow$ & PSNR-N$\uparrow$ & LPIPS$\downarrow$ & Albedo LPIPS$\downarrow$\\
    \midrule
    Degraded (Input) & 1.221 & 9.37 & 1.221 & 0.8821 & 10.31 & 0.2824 & 0.2805\\
    Clean reference (Oracle) & 0.753 & 10.75 & 0.753 & 1.0086 & 10.42 & 0.2760 & 0.2713\\
    \midrule
    FLUX.1 & 1.214 & 9.06 & 1.214 & 0.8582 & 10.26 & 0.2916 & 0.2906\\
    Mage-Flow & 1.259 & 8.88 & 1.259 & 0.8428 & 10.30 & 0.2883 & 0.2864\\
    OmniGen2 & 3.204 & 6.00 & 3.204 & 0.5700 & 10.03 & 0.3051 & 0.3047\\
    Step1X & 1.411 & 8.79 & 1.412 & 0.8289 & 10.20 & 0.2911 & 0.2897\\
    GPT Image-2.5 Flare$^{*}$ & 1.467 & 8.29 & 1.468 & 0.7864 & \underline{10.53} & 0.2825 & 0.2836\\
    Qwen-Image-Edit & 1.150 & \underline{9.29} & 1.150 & \underline{0.8757} & 10.31 & 0.2889 & 0.2883\\
    FireRed & \underline{1.148} & 9.12 & \underline{1.149} & 0.8639 & \textbf{10.77} & \textbf{0.2759} & \underline{0.2775}\\
    \bottomrule
    \rowcolor{supraTableWash}
    \textbf{Ours} & \textbf{1.091} & \textbf{10.16} & \textbf{1.091} & \textbf{0.9588} & 10.36 & \underline{0.2796} & \textbf{0.2773}\\
    \bottomrule
    \end{suprafulltable}
    \end{center}
\end{table}

\end{document}